\documentclass[10pt,twocolumn,letterpaper]{article}

\usepackage{cvpr}
\usepackage{times}
\usepackage{epsfig}
\usepackage{graphicx}
\usepackage{amsmath}
\usepackage{amssymb}

\usepackage{enumitem}
\usepackage{amsmath}
\usepackage{algorithm}
\usepackage[noend]{algpseudocode}
\usepackage[font=small,labelfont=bf]{caption}
\usepackage{graphicx}
\usepackage{subfig}

\usepackage[utf8]{inputenc} 
\usepackage[T1]{fontenc}    
\usepackage{url}            
\usepackage{booktabs}       
\usepackage{amsfonts}       
\usepackage{nicefrac}       
\usepackage{microtype}      
\usepackage{soul}
\usepackage{longtable}
\usepackage{placeins}
\usepackage[table]{xcolor}

\definecolor{Gray}{gray}{0.87}
\newcolumntype{g}{>{\columncolor{Gray}}r}

\usepackage[pagebackref=true,breaklinks=true,letterpaper=true,colorlinks,bookmarks=false]{hyperref}

\cvprfinalcopy 

\begin{document}

\title{Rehearsal-Free Continual Learning over Small Non-I.I.D. Batches}

\author{Vincenzo Lomonaco, Davide Maltoni, Lorenzo Pellegrini\\
University of Bologna\\
Via dell'Università, 50, Cesena, Italy\\
{\tt\small \{vincenzo.lomonaco, davide.maltoni, l.pellegrini\}@unibo.it}
}

\maketitle
\newcommand{\red}[1]{\textcolor{red}{#1}}

\begin{abstract}
Robotic vision is a field where continual learning can play a significant role. An embodied agent operating in a complex environment subject to frequent and unpredictable changes is required to learn and adapt continuously. In the context of object recognition, for example, a robot should be able to learn (without forgetting) objects of never before seen classes as well as improving its recognition capabilities as new instances of already known classes are discovered. Ideally, continual learning should be triggered by the availability of short videos of single objects and performed on-line on on-board hardware with fine-grained updates. In this paper, we introduce a novel continual learning protocol based on the CORe50 benchmark and propose two rehearsal-free continual learning techniques, CWR* and AR1*, that can learn effectively even in the challenging case of nearly 400 small non-i.i.d. incremental batches. In particular, our experiments show that AR1* can outperform other state-of-the-art rehearsal-free techniques by more than 15\% accuracy in some cases, with a very light and constant computational and memory overhead across training batches.
\end{abstract}

\section{Introduction}

Consolidating and preserving past memories while being able to learn new concepts and skills is a well-known challenge for both artificial and biological learning systems, generally acknowledged as the plasticity-stability dilemma \cite{Mermillod2013}. In particular, gradient-based architectures are often skewed towards plasticity and prone to catastrophic forgetting when learning over a stream of non-stationary data \cite{French1999, Robins1995, Kemker2018}.
A simple solution to deal with this issue would be storing all the data, and cyclically re-train the entire model from scratch \cite{Kading2017}. However, this approach is rather impractical when learning from high-dimensional streaming data, especially in highly constrained computational platforms and embedded systems \cite{Lesort2019,Sunderhauf18}.

\begin{figure}%
\centering
\includegraphics[width=\columnwidth]{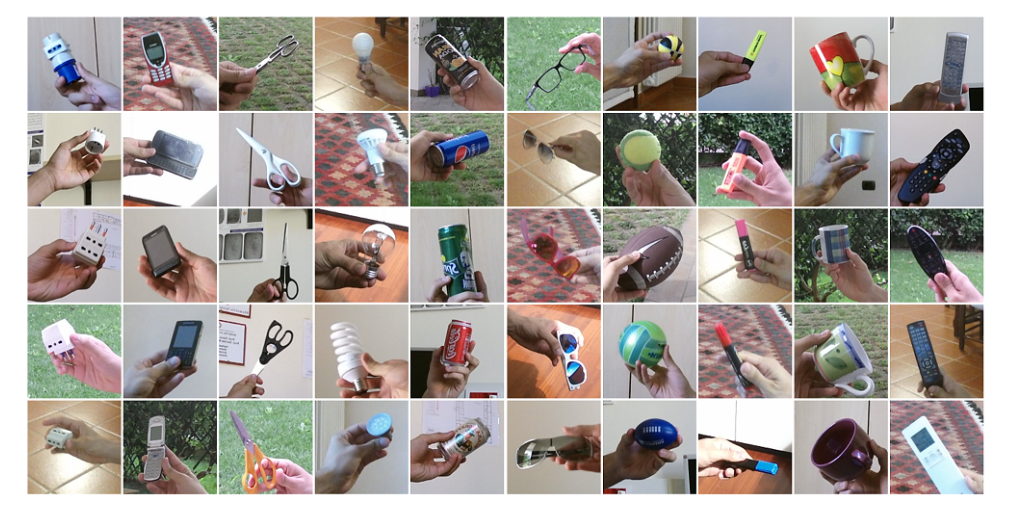}

\caption{Example images of the 50 objects in CORe50, the continual learning video benchmark used in this paper. Each column denotes one of the 10 categories. Classification experiments in this paper are object-based, so each object corresponds to a class.}
\label{fig:core50}
\end{figure}

In recent years, a number of continual learning (CL) strategies have been proposed for deep architectures based on \emph{regularization}, \emph{architectural} or \emph{rehearsal} approaches \cite{Parisi2019, lomonaco2018thesis, Chen2018}. Most of the proposals target a Multi-Task (MT) setting where a sequence of independent and  tasks are encountered over time. However, for many practical applications, such as natural object recognition, a Single-Incremental-Task (SIT) setting may appear more appropriate \cite{maltoni2019}. In the SIT setting, we can distinguish three different cases, based on the training batches composition:
\begin{enumerate}
    \itemsep0.01em 
	\item \textbf{New Instances (NI)}: new training patterns of the same classes become available in subsequent batches with new poses and environment conditions (illumination, background, occlusions, etc.).
    \item \textbf{New Classes (NC)}: new training patterns belonging to different, previously unseen, classes become available in subsequent batches. This is also known as class-incremental learning.
    \item \textbf{New Instances and Classes (NIC)}: new training patterns belonging to both known and new classes become available in subsequent training batches.
To the best of our knowledge, almost no study explicitly addresses the NIC scenario, which we deem as the most natural setting for many applications such as robotics vision, where: i) a large number of small non-i.i.d. training batches are encountered over time; ii) training batches may contain objects already seen before as well as completely new objects. 
\end{enumerate}

Although some researchers pointed out that reducing the size of training batches makes continual learning more challenging \cite{maltoni2019,Hayes2018a,Rebuffi2017}, we still do not know what is the lower bound for the size of training batches and if it is actually feasible to train a system by gradient descent with very small non-i.i.d. incremental batches each containing few images of a single class. It is well known that stochastic gradient descent (SGD) works well with large and i.i.d. mini-batches, but this assumption is difficult to meet. Let us consider a robot that is learning to recognize some objects shown by an operator (one at the time). In an ideal application, when a new object is shown, the robot acquires a short video and immediately updates its knowledge to become able to recognize the new object. The frames extracted from the video would constitute one or more small mini-batches containing highly correlated patterns from a single class: a rather challenging setting to face with standard SGD-based optimization techniques.

Some rehearsal-based techniques have been proposed in order to mitigate this problem: by maintaining some representative patterns from past experiences, new frames can be interleaved with past ones in each mini-batch. However, this involves extra memory (to store the past data) and computation (due to an higher number of forward/backward steps): in this work we ask ourselves weather continual learning over small non-i.i.d. batches \emph{is feasible without rehearsal}.

The contributions of this paper can be summarized as follows:

\begin{itemize}
    \itemsep0.01em 
    \item we propose two rehearsal-free continual learning strategies, CWR* and AR1*, as extensions of the CWR+ and AR1 strategies originally proposed for the NC scenario in \cite{maltoni2019}, making them agnostic to the batches composition.
    \item we show that replacing Batch Normalization with Batch Renormalization \cite{loffe2017} allows SGD to continually learn even in the challenging case of very small and non i.i.d. batches.
    \item we introduce two different approaches, namely \emph{depthwise layer freezing} and \emph{weight constraining by learning rate modulation} aimed at reducing storage/computation of existing continual learning techniques without hindering accuracy.
    \item we design and openly release at \url{https://vlomonaco.github.io/core50} a new NIC protocol based on CORe50 \cite{pmlr-v78-lomonaco17a} to explicitly address non-i.i.d continual learning scenarios (with 79, 196 and 391 training batches). To the best of our knowledge, this is one of the first attempts to scale continual learning techniques over hundreds of small training batches with real-world highly correlated images.
    \item we run several experiments to evaluate the proposed strategies (CWR* and AR1*) and to compare them with two baselines and three state-of-the-art rehearsal-free approaches (such as EWC \cite{Kirkpatrick2016}, LWF \cite{Li2016} and DSLDA \cite{Hayes2019}), also in terms of computation and memory efficiency.
\end{itemize}

\section{Continual Learning Strategies}
\label{sec:strats}

In \cite{maltoni2019} it was showed that a simple approach like CWR+, where the fully connected layer is implemented as a double memory, is quite effective to control forgetting in the SIT - NC scenario.
However, after the first training batch, CWR+ freezes all the layers except the last one, thus losing the benefit of an incremental adaptation of the underlying representation. AR1 \cite{maltoni2019} was then proposed to extend CWR+ by enabling end-to-end continual training throughout the entire network; to this purpose the Synaptic Intelligence \cite{Zenke2017} regularization approach (similar to EWC \cite{Kirkpatrick2016}) is adopted to constrain the change of critical weights.
In the following subsections we:
\begin{enumerate}
    \itemsep0.01em 
	\item adapt CWR+ to the NIC scenario, thus making it able to reload past weights for already known classes and to adapt them with weighted contributions from different batches. As AR1 incorporates CWR+ in its main algorithm, this modification will result in two continual learning strategies hereby denoted as CWR* and AR1* (Section \ref{subsec:cwr*}).
	\item show that in a complex scenario with small and non i.i.d. batches, Batch Normalization layers thwart the continual learning process and replacing them with Batch Renormalization \cite{loffe2017} can effectively tackle this problem (Section \ref{subsec:brn}).
	\item propose a selective weight freeze for the CNN models adopting Depth-Wise Separable Convolutions (Section \ref{subsec:freeze}).
	\item reduce the computational and storage complexity of AR1 (and in general of EWC like approaches), by introducing an alternative way to implement weights update starting from the Fisher matrix (Section \ref{subsec:mod}).
\end{enumerate}

While 1. is specific to CWR+, 2., 3. and 4. can be applied to several other CL approaches as well.


\subsection{From CWR+ to CWR*}
\label{subsec:cwr*}

CWR+, whose pseudo-code is reported in Algorithm 2 of \cite{maltoni2019} and in the supplementary materials of this work, maintains two sets of weights for the output classification layer: $cw$ are the consolidated weights used for inference and $tw$ the temporary weights used for training; $cw$ are initialized to 0 before the first batch and then iteratively updated, while $tw$ are reset to 0 before each training batch.

In Algorithm \ref{algo:cwr*}, we propose an extension of CWR+ called CWR* which works both under NC and NIC update type; in particular, under NIC the coming batches include patterns of both new and already encountered classes. For already known classes, instead of resetting weights to 0, we reload the consolidated weights (see line 7). Furthermore, in the consolidation step (line 13) a weighted sum is now used: the first term represents the weight of the past and the second term is the contribution from the current training batch. The weight $wpast_j$ used for the first term is proportional to the ratio $\frac{past_j}{cur_j}$, where $past_j$ is the total number of patterns of class $j$ encountered in past batches whereas $cur_j$ is their count in the current batch. In case of a large number of small non-i.i.d. training batches, the weight for the most recent batches may be too low thus hindering the learning process. In order to avoid this, a square root is used in order to smooth the final value of $wpast_j$.

\begin{algorithm}[h]
\captionsetup{font=small}
\caption{CWR* pseudocode: $\bar{\Theta}$ are the class-shared parameters of the representation layers; the notation  $cw[j]$ / $tw[j]$ is used to denote the groups of consolidated / temporary weights corresponding to class $j$. Note that this version continues to work under NC, which is seen here a special case of NIC; in fact, since in NC the classes in the current batch were never encountered before, the step at line 7 loads 0 values for classes in $B_i$ because $cw_j$ were initialized to 0 and in the consolidation step (line 13) $wpast_j$ values are always 0.}
\label{algo:cwr*}
\begin{algorithmic}[1]
\footnotesize
\Procedure{CWR*}{}
\State $cw=0$
\State $past=0$
\State $\text{init } \bar{\Theta} \text{ random or from pre-trained model (e.g. on ImageNet)}$
\State \textbf{for each} $\text{training batch } B_i$:
\State \ \ \ expand output layer with neurons for the new classes in $B_i$ \phantom \ \phantom \ \phantom \ \phantom \ \phantom \ \phantom \ \phantom \ \phantom \ \phantom \ \phantom \ never seen before
\State \ \ \ \ $tw[j]=
    \begin{cases}
      cw[j], & \text{if class } j \text{ in } B_i\\
      0, & \text{otherwise}
    \end{cases}$
\State \ \ \ \ train the model with SGD on the $s_i$ classes of $B_i$:
\State \ \ \ \ \ \ \ \ \textbf{if} $B_i=B_1$ learn both $\bar{\Theta}$ and $tw$
\State \ \ \ \ \ \ \ \ \textbf{else} learn $tw$ while keeping $\bar{\Theta}$ fixed
\State \ \ \ \ \textbf{for each} class $j$ in $B_i$:
\State \ \ \ \ \ \ \ \ $wpast_j = \sqrt{\frac{past_j}{cur_j}}$, \text{where } $cur_j$ is the number of patterns \phantom \ \phantom \ \phantom \ \phantom \ \phantom \ \phantom \ \phantom \ \phantom \ \phantom \ \phantom \ \ \ \ \ of class $j$ in $B_i$
\State \ \ \ \ \ \ \ \ $cw[j]=\frac{cw[j] \cdot wpast_j + (tw[j]-avg(tw))}{wpast_j + 1}$
\State \ \ \ \ \ \ \ \ $past_j = past_j + cur_j$
\State \ \ \ \ test the model by using $\bar{\Theta}$ and $cw$
\EndProcedure
\end{algorithmic}
\end{algorithm}
	

\subsection{Replacing Batch Normalization with Batch Renormalization}
\label{subsec:brn}

Batch Normalization (BN) \cite{loffe2015} is widely used in modern deep neural networks to control internal covariate shift thus making learning faster and more robust. In BN the mini-batch moments (i.e., mean $\mu_{mb}$ and variance $\sigma_{mb}^2$) are used to normalize the input values $x_i$ as:
\begin{equation}
  \hat{x_i} = \frac{x_i - \mu_{mb}}{\sqrt{\sigma_{mb}^2} + \epsilon}
\end{equation}
where $\epsilon$ is a small constant added for numerical stability, and the normalization is per-channel. However, if mini-batches are small and/or non i.i.d. the mini-batch moments are not stable and BN can fail. A natural solution to reduce the moment fluctuations would be replacing $\mu_{mb}$, $\sigma_{mb}^2$ with global values $\mu$, $\sigma$  computed as moving averages over an initial (large-enough) training batch. After all, this is the standard approach when switching from training to inference. However, as argued in \cite{loffe2015}, using moving averages to perform the normalization during training does not produce the desired effects since gradient optimization and the normalization counteract each other, possibly leading the model to diverge.

Batch Renormalization (BRN) was proposed in \cite{loffe2017} to deal with small and non i.i.d. mini-batches. In BRN the normalization takes place as follows:

\begin{equation}
  \hat{x_i} = \frac{x_i - \mu_{mb}}{\sigma_{mb}} \cdot r + d
  \label{eq:first}
\end{equation}

\begin{equation}
  r=\frac{\sigma_{mb}}{\sigma}, \quad d= \frac{\mu_{mb}-\mu}{\sigma}
\end{equation}

\noindent where $\mu$, $\sigma$ are computed as moving averages during training. By expanding $r$ and $d$ in the equation \ref{eq:first}, we obtain $\hat{x_i} = \frac{x_i - \mu}{\sigma}$ which clearly highlights the dependency on the global moments. A further step is suggested in \cite{loffe2017} to clip $r$ in $[\frac{1}{r_{max}}, r_{max}]$ and $d$ in $[-d_{max}, d_{max}]$. It is worth noting that when $r=1$ and $d=0$, then BRN$\equiv$BN; hence, by properly setting $r_{max}$ and $d_{max}$ the behavior of BRN can be moved from a pure BN to a more stable normalization based on global statistics. In practice, the author of \cite{loffe2017} recommend to perform an initial stage by keeping $r_{max}=1$, $d_{max}=0$ in order to stabilize the moving averages $\mu$, $\sigma$ and then progressively increasing $r_{max}$ and $d_{max}$ to 3 and 5, respectively.

Continual learning over small batches is an emblematic case of small and non i.i.d. minibatches. For example, in NICv2-391 (introduced in Section \ref{sec:nicv2}) each training batch includes 300 patterns from a single class, and even using a mini-batch size of 300 (the full batch) patterns remain strongly correlated. Our first attempts to learn continuously over a so long sequence of one-class batches were totally unsatisfactory. Even for the most accurate strategies (e.g., AR1*) accuracy slightly increased in the first batches from 13\% to 15\% but then remained steady and lower than 16-17\%. We initially though that the reason were the single-class mini-batches, making the problem a sort of one-class classification with no negative examples. However, upon replacement of BN with BRN and a proper parametrization, we were able to continuously learn over small batches with unexpected efficacy (see Section \ref{sec:result} for optimal parametrization and results). 

\subsection{Depthwise Layer Freezing}
\label{subsec:freeze}

Depth-Wise Separable Convolutions (DWSC) are quite popular nowadays in many successful CNN architectures such as MobileNet \cite{Howard2012,Sandler2018}, Xception \cite{Chollet2017}, EfficientNet \cite{Tan2019}. Classical filters in CNN are shaped as 3D volumes. For example, a 5$\times$5$\times$32 filter spans a spatial neighborhood of 5$\times$5 along 32 feature maps; on the contrary, in DWSC we first perform 32 5$\times$5$\times$1 spatial convolutions (an independent convolution on each feature maps) and then combine results with a 1$\times$1$\times$32 filter working as a feature map pooler. Advantages in terms of computation and weight reduction have been pointed out by several researchers. 

Inspired by previous finding with Hierarchical Temporal Memories \cite{Rehn2014, Maltoni2016-icpr}  where gradient descent by HSR only affects coincidence pooling, here we propose to fine-tune DWSC architectures by freezing depthwise spatial filters and leaving pointwise poolers free to learn. We speculate that modifying a spatial filter (i.e. the way a local neighborhood is processed) can be detrimental in terms of forgetting during continual learning, because it alters the semantics of what upper layers have already learned; on the other hand, feature map pooling, which can be seen as a way to promote feature invariance, is less prone to concept drifts.

\begin{figure}%
\centering
\includegraphics[width=0.8\columnwidth]{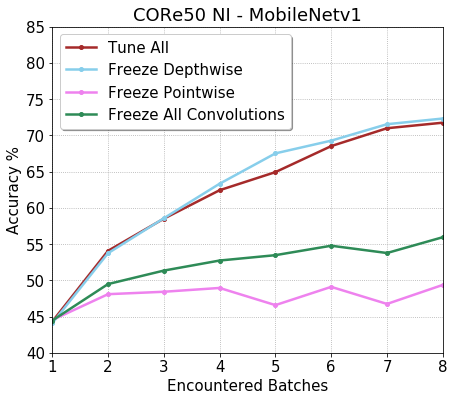}

\caption{Continual Learning on CORe50, SIT – NI scenario. In the NI scenario all the 50 classes are discovered in the first batch and successive training batches provide new instances of the already known classes; however, since past instances are not retained, the incremental process is prone to forgetting. In this experiment a MobileNet (pre-trained on ImageNet) is incrementally tuned (naïve strategy) over 8 batches with different weight freeze strategies. Each curve is averaged over 10 runs where the batch order is randomly permuted.}
\label{fig:NI}
\end{figure}

A simple experiment is illustrated in figure \ref{fig:NI}, where a MobileNet is incrementally fine-tuned along the 8 learning batches of CORe50, SIT – NI scenario \cite{pmlr-v78-lomonaco17a}. Here, no specific measure is put in place to control forgetting except early stopping the gradient descent after 1 epoch (naïve strategy). The four curves denote the classification accuracy when: i) all the weights are tuned; ii) weights of depthwise convolution layers are frozen; iii) weights of pointwise convolution layers are frozen; iv) weights of all convolution layers are frozen. Note that weights of fully connected layers (e.g. output layer) are never frozen. The proposed strategy (case ii) achieves the best result and, with respect to a full tuning, allows skipping some gradient computations and can reduce the amount of memory used to store weight associated data\footnote{In per-weight adaptive learning rate methods (such as Adam \cite{Kingma2014}) extra values (i.e. running averages) need to be stored for each “free” weight. Further, if a regularization method based on Fisher matrix is used (such as EWC \cite{Kirkpatrick2016}) we need to store the optimal value for previous tasks and the Fisher value for each weight.}. The complementary strategy (case iii) is the worst one, thus confirming that altering spatial filters has a strong impact in terms of forgetting.

\begin{figure*}[t]
    \centering
  \includegraphics[width=0.9\textwidth]{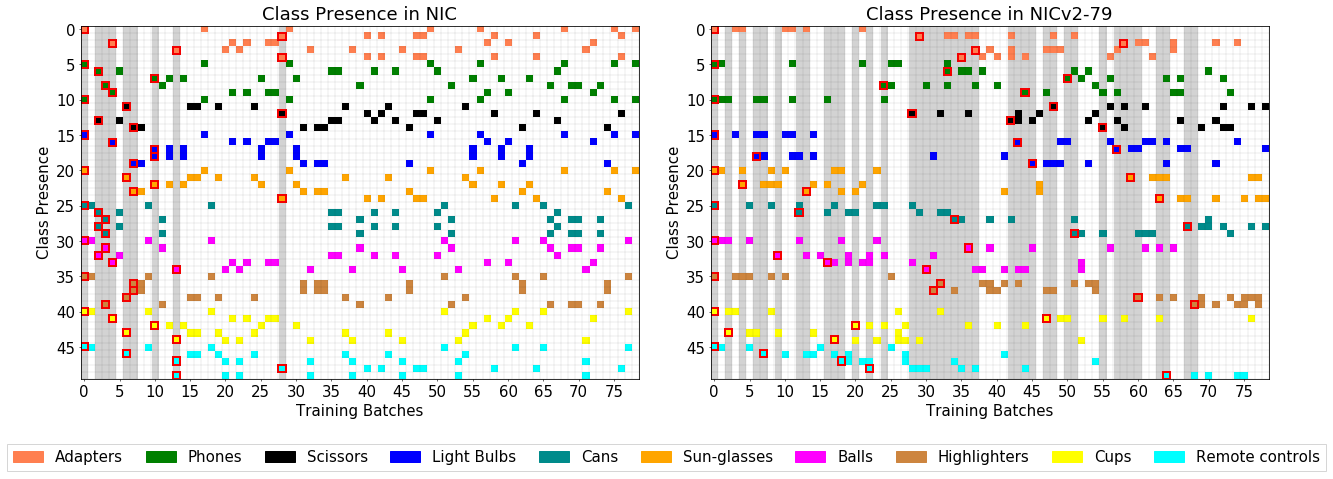}
  \caption{Classes encountered over time in the first run of NIC (left) and NICv2-79 (right). Each row denotes a class; colors are used to group the 50 classes in the 10 categories. Each column denotes a training batch. A colored block in a (row, column) cell is used to indicate that at least one training session of the row class is present in the column batch. Each row contains a maximum of 8 colored blocks, because each class has 8 training sessions in the training set (the remaining 3 sessions are left out for the test set). The red-framed cells denote the first introduction of a class. Gray vertical bands highlight batches where at least one class is seen for the first time. Class presentation ordering for the first run of NICv2-196 and NICv2-391 is also reported in Figure \ref{fig:nic196} and Figure \ref{fig:nic391} of the appendix.}
  \label{fig:nicv2}
\end{figure*}

\subsection{Weight Constraining by Learning Rate Modulation}
\label{subsec:mod}

The Elastic Weight Consolidation (EWC) approach \cite{Kirkpatrick2016} tries to control forgetting by selectively constraining the model weights which are deemed to be important for the previous tasks. To this purpose, in a classification approach, a regularization term is added to the conventional cross-entropy loss, where the weights $\theta_k$ of the model are pulled back to their optimal value $\theta_k^*$ with a strength $F_k$ proportional to their importance for the past: 

\begin{equation}
L=L_{cross}(\cdot) + \frac{\lambda}{2} \cdot \sum_{k} F_k \cdot (\theta_k - \theta_k^*)^2.
\label{eq:loss}
\end{equation}

Synaptic Intelligence (SI) \cite{Zenke2017} is a lightweight variant of EWC where, instead of updating the Fisher matrix $F$ at the end of each batch\footnote{In this paper, for the EWC and AR1 implementations we use a single Fisher matrix updated over time, following the approach described in \cite{maltoni2019}.}, $F_k$ are obtained by integrating the loss over the weight trajectories exploiting information already available during gradient descent. For both approaches, the weight update rule corresponding to equation \ref{eq:loss} is:

\begin{equation}
\theta^{'}_k = \theta_k - \eta \cdot \frac{\partial L_{cross}(\cdot)}{\partial\theta_k} - \eta \cdot F_k \cdot (\theta_k - \theta_k^*)
\label{eq:update}
\end{equation}

\noindent where $\eta$ is the learning rate. This equation has two drawbacks. Firstly, the value of $\lambda$ must be carefully calibrated: in fact, if its value is too high the optimal value of some parameters could be overshoot, leading to divergence (see discussion in Section 2 of \cite{maltoni2019}). Secondly, two copies of all model weights must be maintained to store both $\theta_k$ and $\theta_k^*$, leading to double memory consumption for each weight.
To overcome the above problems, we propose to replace the update rule of equation \ref{eq:update} with:

\begin{equation}
 \theta^{'}_k = \theta_k - \eta \cdot (1 - \frac{F_k}{max_F}) \cdot \frac{\partial L_{cross}(\cdot)}{\partial\theta_k}
\end{equation}

\noindent where $max_F$ is the maximum value for weight importance (we clip to $max_F$ the $F_k$ values larger than $max_F$). Basically, the learning rate is reduced to 0 (i.e., complete freezing) for weights of highest importance ($F_k=max_F$) and maintained to $\eta$ for weights whose $F_k=0$.  It is worth noting that these two updated rules work differently: the former still moves weights with high $F_k$ in direction opposite to the gradient and then makes a step in direction of the past (optimal) values; the latter tends to completely freeze weights with high $F_k$. However, in our experiments with AR1 the two approaches lead to similar results, and therefore the second one is preferable since it solves the aforementioned drawbacks.

\section{CORe50 NICv2}
\label{sec:nicv2}

CORe50 \cite{pmlr-v78-lomonaco17a} was specifically designed as an object recognition video benchmark for continual learning. It consists of 164,866 128$\times$128 images of 50 domestic objects belonging to 10 categories (see Figure \ref{fig:core50}); for each object the dataset includes 11 video sessions ($\sim$300 frames recorded with a Kinect 2 at 20 fps) characterized by relevant variations in terms of lighting, background, pose and occlusions. The egocentric vision of hand-held objects allows emulating a scenario where a robot has to incrementally learn to recognize objects while manipulating them. Objects are presented to the robot by a human operator who can also provide the labels, thus enabling a supervised classification (such an applicative scenario is well described in \cite{pasqualevideo, Pasquale2015a,Pasquale2019a}).

\begin{table*}[t]
  \caption{Batch number and composition in NIC and NICv2.}
  \label{tab:nic}
  \renewcommand{\arraystretch}{0.95}
  \centering
  \begin{tabular}{p{2cm}p{1.5cm}p{1.3cm}p{1.3cm}p{1.3cm}p{1.3cm}}
    \toprule
    Protocol & \# batches  & \multicolumn{2}{c}{Initial batch} & \multicolumn{2}{c}{Incremental Batches}\\
    \cmidrule{3-6}
     &  & \# Classes & \# Images & \# Classes & \# Images\\
    \midrule
    NIC & 79  & 10 & 3,000 & 5 & 1,500\\
    NICv2-79 & 79  & 10 & 3,000 & 5 & 1,500\\
    NICv2-196 & 196  & 10 & 3,000 & 2 & 600\\
    NICv2-391 & 391  & 10 & 3,000 & 1 & 300\\
    \bottomrule
  \end{tabular}
\end{table*}

A NIC protocol was initially introduced for CORe50 \cite{pmlr-v78-lomonaco17a} where the first training batch contains 10 classes ($\sim$3,000 images) and each of the subsequent 78 incremental batches includes about 1,500 images of 5 classes. However, as shown in Figure \ref{fig:nicv2} (left), the random generation procedure used in \cite{pmlr-v78-lomonaco17a} produced a sequence where almost all the classes are introduced in the first 10-15 batches making this protocol very close to an NI scenario.

To make the benchmark more challenging and closer to a real application where new objects can be discovered also later in time, we propose a new three-way protocol (denoted as NICv2) where classes first introduction is more balanced over the training batches (see Figure \ref{fig:nicv2}, right) and the batch size is progressively reduced, leading to a higher number of fine-grained updates (see Table \ref{tab:nic}). In particular, in NICv2-391 each of the 390 incremental batches includes only one training session ($\sim$300 images) of a single class. The pseudo-code used to generate the NICv2 protocol is reported in Algorithm \ref{algo:nicv2} in the appendix.

The test set used for NICv2 is the default test set shared by all the CORe50 protocols \cite{pmlr-v78-lomonaco17a}; it includes 3 sessions for each class, with null intersection with training batches. Actually, in order to speed up the large number of evaluations (which requires one evaluation after each training batch, repeated for 10 runs) we sub-sampled the test set by selecting 1 frame every second (from the original 20 fps). Because of the high correlation among successive frames in the sequences, such a strong sub-sampling is not reducing the test set variability and the accuracy results on the original and the down sampled version are very close. We made available at \url{https://vlomonaco.github.io/core50} all the file lists of the new NICv2 protocols along with the down-sampled test set.

\section{Experimental Results}
\label{sec:result}

We run several experiments on CORe50 NICv2, to validate the approaches introduced in Section \ref{sec:strats} and to compare them with a naïve baseline and three state-of-the-art rehearsal-free approaches. In particular, for all the experiments, the following techniques have been considered:

\begin{itemize}
    \itemsep0.01em 
    \item CWR*: the extension of CWR+ introduced in Section \ref{sec:strats}.
    \item AR1*: the approach introduced in \cite{maltoni2019}, here implemented by replacing CWR+ with CWR* and by adopting the weight constraining by learning rate modulation introduced in Section \ref{subsec:freeze}.
    \item Naïve: a baseline technique where we simply continue gradient descent along the training batches and the only measure to control forgetting is early stopping.
	\item EWC and LFW: the techniques originally introduced in \cite{Kirkpatrick2016} and \cite{Li2016} and adapted to continual learning in SIT scenario as detailed in \cite{maltoni2019}.
	\item DSLDA: the strategy recently proposed in \cite{Hayes2019}, where an on-line extension of the Linear Discriminant Analysis (LDA) classifier \cite{Rao1948} is trained on the top of a fixed deep learning feature extractor. DSLDA obtained state-of-the art accuracy on CORe50 (10 categories setting) \cite{Hayes2019}, even outperforming rehearsal based techniques such as ICARL \cite{Rebuffi2017} and ExStream \cite{Hayes2018a}.
    \item Cumulative: this is a sort of upper bound in terms of accuracy because the model is trained on the union of the current batch and all the past data.
\end{itemize}

For all the experiments we used a MobileNet v1 \cite{Howard2012} with: $width\ multiplier = 1$, $resolution\ multiplier = 0.5$ (input 128 $\times$ 128), pre-trained on ImageNet. MobileNet architectures provide a good tradeoff in terms of accuracy/efficiency and, in our opinion, are well suited for porting continual learning at the edge. 

For all the above techniques the MobileNet v1 architecture was modified by replacing the 27 Batch Normalization layers 
with corresponding Batch Renormalization layers and using (for training) a mini-batch size of 128 patterns. We used Batch Renormalization implementation for Caffe \cite{Jia2014} made available in \cite{brncode}. This modification improves accuracy of all the methods, making CWR* and AR1* able to learn also in the case of 391 single class batches. 
Batch Renormalization hyperparameters and their schedule have been experimentally set as follows:

\begin{itemize}
\itemsep0.01em 
\item \textbf{Batch 1}: for the first 48 iterations we keep $r_{max}=1$, $d_{max}=0$ to startup the global moments; then, we progressively move $r_{max}$ to 3 and $d_{max}$ to 5 (as suggested in \cite{loffe2017}). The weight of the past when updating the moving averages was set to 0.99 (as suggested for ($1-\alpha$) in \cite{loffe2017}).
\item \textbf{Subsequent batches}: global moments computed on batch 1 are inherited by batch 2 and slowly updated across the batch sequence. In this case we noted that continual learning over small non i.i.d. batches benefits from more stable moments, and therefore the weight of the past for updating moving averages was set to 0.9999. Here we have no startup phase for the global moment so the values of $r_{max}$ and $d_{max}$ are kept fixed across all the iterations of the batches. While using the suggested values of $r_{max}=3$ and $d_{max}=5$ still works, we noted that reducing them (i.e. relaxing batch renormalization constrains) brings some befits. More details about the experiments and the hyperparameters used are provided in the appendix.
\end{itemize}

\begin{figure*}[h]
\center
\noindent\includegraphics[width=0.93\textwidth]{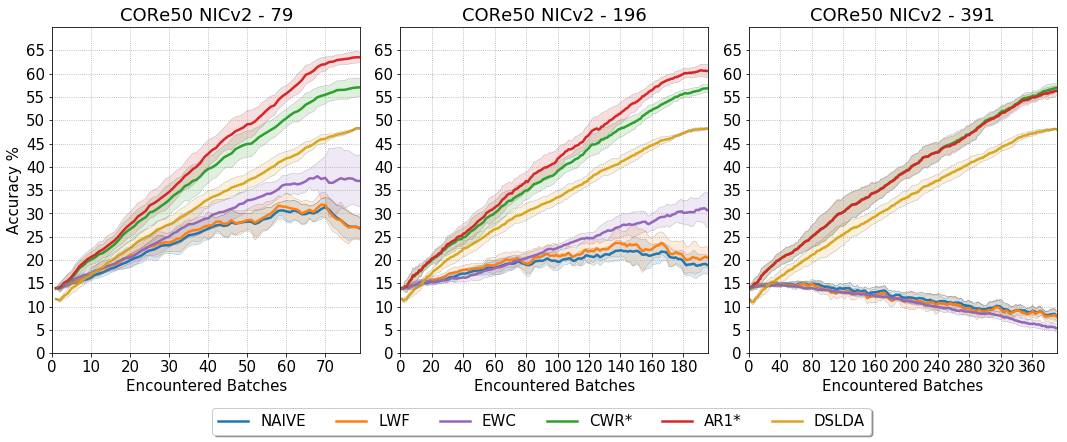} 
\caption{Continual learning accuracy over NICv2-79, NICv2-196 and NICv2-391. Each experiment was averaged on 10 runs. Colored areas represent the standard deviation of each curve. Accuracy performance for the cumulative upper bound (trained on the union of all batches), not reported for visual convenience, is $\sim$85\%. Results in tabular form are made available for download at \url{https://vlomonaco.github.io/core50}.}
\label{fig:results}
\end{figure*}

For all the techniques we also applied depthwise layer freezing (as introduced in Section \ref{subsec:freeze}) starting from Batch 2. This can be simply implemented by setting learning rate to 0 for the 14 non pointwise convolution layers (13 depthwise + 1 3D) in MobileNet v1 architecture. While in NICv2 experiments this had a negligible impact on the accuracy, we found it can be advantageous in other scenarios (see NI curves in Figure \ref{fig:NI}) and, in general, this reduces computations/storage during SGD (less gradient calculations, lower memory to accumulate per weight extra-data, etc.).

Figure \ref{fig:results} shows the results of our experiments on NICv2-79, NICv2-196 and NICv2-391. The curves are averaged over 10 runs where the training batch order is randomly shuffled. Hyperparameters of the methods have been coarsely tuned (i.e., without any systematic grid search) on run 0 and then kept fixed for the other 9 runs.
It can be noted that CWR* and AR1* show a very good learning trend across training batches, with only a minor drop in accuracy when the batch granularity decreases. The accuracy near linearly increases for most of the batches and slows down in the final part of the sequences; we believe this is not caused by the saturation of learning capabilities but is more likely due to the absence of example of new classes in the final part of the sequences (see Figure \ref{sec:nicv2}b). Standard deviation across runs is also quite small denoting a good stability. Naïve, LWF and EWC exhibit fair performance on 79 batches but their efficacy significantly decreases with 196 batches and are not able to learn in the most challenging case of 391 single-class batches. DSLDA accuracy is quite good and stable but remains lower than CWR* and AR1* in all the three settings. The advantage of AR1* over CWR* (due to the extra freedom to improve the representation) reduces as the batch size decreases and is null for 391 batches. We speculate that, in this case, the gradient steps induced by small and highly non i.i.d. mini-batches tend to overfit the mini-batches themselves with no improvement in term of generalization.

\begin{figure*}[h]
\center
\noindent\includegraphics[width=0.92\textwidth]{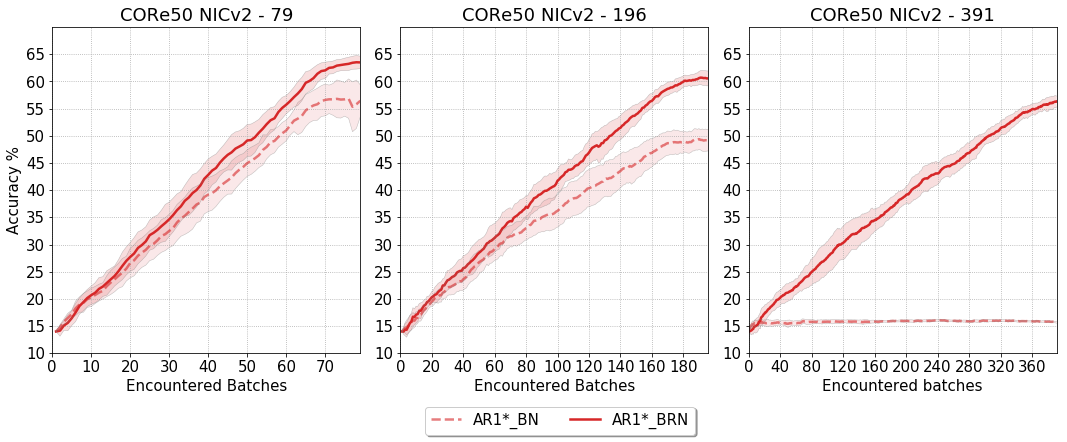} 
\caption{Comparing AR1* accuracy results in the Batch Normalization vs Batch Renormalization configurations.}
\label{fig:bn_vs_brn}
\end{figure*}

Figure \ref{fig:bn_vs_brn} compares AR1* accuracy in the configuration with Batch Normalization and Batch Renormalization. It is evident that for 391 batches Batch Normalization heavily hurt the learning capabilities. However, it is worth noting that Batch Renormalization brings some advantages to continual learning even when using larger batches that may include patterns from more than one class. 

\begin{table}[h]
  \caption{Total run time (in minutes, for both training and test), memory overhead (in terms of maximum data storage for rehearsal and number of additional trainable parameters introduced) for each strategy on the NICv2-391 protocol. Please note that: \emph{(i)} all the rehearsal-free strategy hereby listed have a \emph{constant} memory / computational overhead which is fixed and independent from the number of training batches processed; \emph{(ii)} the Cumulative metrics are computed considering a re-training from scratch after each incremental batch.}
  \label{tab:efficiency}
  \centering
  \small
  \begin{tabular}{p{0.9cm}p{2cm}p{1.5cm}p{1.7cm}}
    \toprule
    & & \multicolumn{2}{c}{Memory Overhead} \\
    Strat. & Run Time (m) & Data (MB) & Params (MB)\\
    \midrule
    CWR* & 21,4 & 0 & 0,2\\
    Naive & 25,6 & 0 & 0\\
    LWF & 27,8 & 0 & 0\\
    EWC & 31,2 & 0 & 24,4\\
    \textbf{AR1*} & \textbf{39,9} & \textbf{0} & \textbf{12,2}\\
    DSLDA & 79,1 & 0 & 0,2\\
    Cumul. & 2826,2 & 4712,3 & 0\\
    \bottomrule
  \end{tabular}
\end{table}

In order to better understand and compare the performance of the proposed continual learning strategies, in Table \ref{tab:efficiency} we also report the total run time, the maximum external memory size (where patterns from previous batches are stored) and the number of additional trainable parameters introduced while learning across the NICv2-391 batches. All the metrics are averaged across 10 runs.

Rehearsal-free approaches show a remarkable advantage w.r.t. the cumulative upper bound (where the model is re-trained from scratch after each
incremental batch on the cumulated data), both in terms of speed-up and in terms of total memory overhead. Among them, AR1* shows the best trade-off between accuracy and efficiency with about 40 minutes to complete the run and a fixed memory overhead of only 12.4 MB for handling the additional parameters of the learning rate modulation introduced in Section \ref{subsec:mod}. We would also underline that the current Synaptic Intelligence implementation embedded in AR1* is not optimized (gradient is recomputed in python outside the Caffe framework) without exploiting the data already available from SGD. We believe that upon proper optimization, AR1* efficiency can be very close to Naive one.

Finally, it is worth noting that the advantage of weight constraining by learning rate modulation (introduced in Section \ref{subsec:mod}) for AR1* is negligible in terms of accuracy (less than 0.1\% average improvement in NICv2-79) but relevant in terms of per weight storage since we do not need to store about 3,2 millions $\theta_k^*$ values.


\section{Conclusions}
\label{sec:conclusion}

In this paper, we showed that rehearsal-free continual learning techniques can learn over long sequences of small and highly correlated batches, even in the challenging case of one class at a time. In fact, CWR* and AR1* displayed a good (near linear) learning trend across the training batches and proved to be very robust even with small one-class batches. On the other hand, well known CL techniques such as EWC and LWF were not able to learn effectively in our experiments. We speculate that: (i) a regularization technique alone is not effective to protect important weights in the upper levels when dealing with a large number of small batches; (ii) learning the upper layer(s) ``in isolation'', as CWR* and AR1* do, is very important for continual learning, especially in SIT setting. DSLDA, that recently achieved state-of-art accuracy on some continual learning benckmarks, performed quite well in our experiments, but its accuracy and efficiency are lower than CWR* and AR1*. 

Of course, other continual learning approaches should be considered in the future for a more comprehensive analysis. For example, here we did not consider rehearsal based approaches such as ICARL \cite{Rebuffi2017} and GEM \cite{Lopez-paz2017} because, even if the use of an external memory to store past data may simplify the task, it brings drawbacks in terms of extra storage/computations. Actually, some preliminary comparisons of CWR+, AR1 and DSLDA with rehearsal-based approaches have been reported in \cite{maltoni2019} and \cite{Hayes2019} for CORe50 (NC scenario) showing that the proposed rehearsal-free approaches are still competitive when a moderate number of patterns is maintained in the external memory by ICARL and GEM (2,500-4,500 training images).
Another interesting technique, reporting good results on CORe50, is the Dual-Memory Recurrent Self-Organization proposed in \cite{Parisi2018a}: however, results included in that work are not directly comparable with our achievements because the aforementioned approach also exploits the temporal dimension of CORe50 videos (by using temporal windows instead of single frames).

The top accuracy reached by AR1* at the end of the training sequence is in the range 55-65\% depending on the batch granularity, and the gap w.r.t. cumulative training ($\sim$85\%) exploiting all the data at one time is quite relevant ($>$20\%). In the future, we would like to improve the proposed CL techniques to reduce this gap as much as possible. Pseudo-rehearsal, i.e., generating past data without explicit storage, is the main path we intend to explore. Finally, porting continual learning at the edge, i.e. running end-to-end training algorithms on light architectures with neither remote server support nor on-board GPUs, is another topic of interest for us. In the near future, we plan to release a CWR*/AR1* embedded implementation for smart-phones devices and embedded robotics platforms.

{\small
\bibliographystyle{ieee_fullname}
\bibliography{library}
}

\appendix

In our experiments we used a customized version of the Caffe \cite{Jia2014} framework. More details can be found in the accompanying project source code.
\section{Implementation and Experiments Details}

For each of the proposed scenarios and strategies, a test accuracy curve was obtained by averaging over 10 different runs. Each run differs from the others by the order of the encountered batches. See Algorithm \ref{algo:nicv2} for more details about the NICv2 protocol.
All our experiments were executed in a ``Ubuntu 16.04'' Docker environment using a single GPU. See table \ref{table:hardware} for more details of the host setup.
In the following subsections we report the original CWR+ algorithm as well as the procedure for generating the proposed NICv2 protocol.

\subsection{Original CWR+ Strategy}
The original CWR+ strategy pseudocode is reported for completeness in Algorithm \ref{algo:cwr+}.

The CWR+ strategy was originally targeted at the NC scenario and is not well suited for NI and NIC scenarios due to the fact that previous weights of the CWR layer are overwritten in successive batches (see line 11). This is not an issue in the NC scenario, where each class is encountered in exactly one batch, because class-specific weights are initialized once and are never updated again.

Our proposed strategy CWR* addresses this issue by updating previously learned weights instead of overwriting them.

\subsection{NICv2 Protocol Generation}
The NICv2 protocol was conceived with the idea of balancing the classes first introduction over the training batches. In contrast, in the original NIC protocol classes are introduced in the initial training batches with the direct consequence that the runs generated by NIC end up being too similar to a New Instances (NI) scenario for the remaining batches. Figure 2, in the paper, clearly shows this issue, making it clear that for the vast majority of the scenario no new classes are encountered.

The NICv2 protocol addresses this issue by forcing the classes first introduction to be evenly distributed across the batches thus producing runs that are both more challenging and realistic. We believe that this scenario can be used as a theoretical benchmark for the study of new continual learning strategies to be employed in the robotics and learning at the edge fields. In general, the NICv2 protocol is able to recreate realistic scenarios in which no general assumption can be made on the order and distribution in time in which new classes may be introduced.

\begin{figure*}
    \centering
    \includegraphics[width=1.0\linewidth]{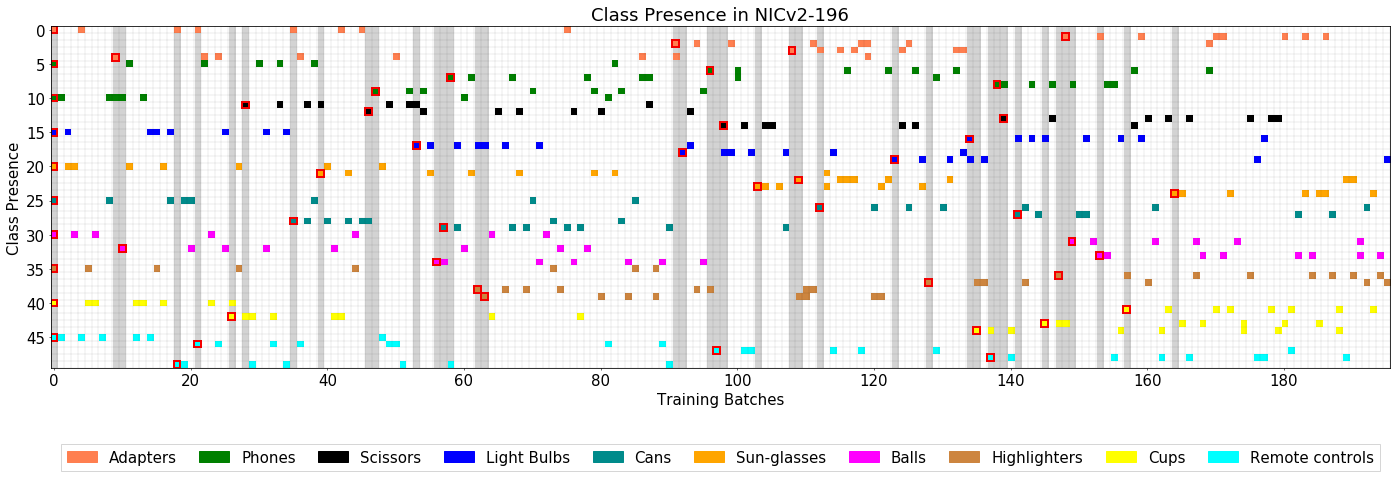}
    \caption{Classes encountered over time in the first run NICv2-196. Each row denotes a class; colors are used to group the 50 classes in the 10 categories.  Each column denotes a training batch. A colored block in a (row, column) cell is used to indicate that at least one training session of the row class is present in the column batch.}
    \label{fig:nic196}
\end{figure*}

\begin{figure*}
    \centering
    \includegraphics[width=1.0\linewidth]{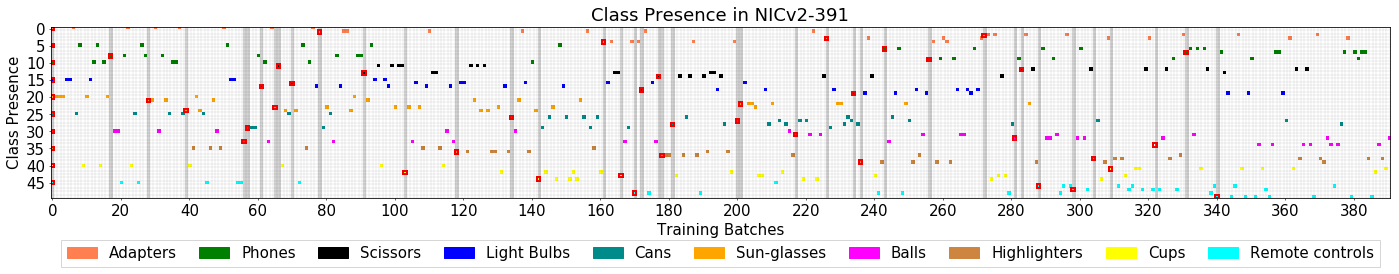}
    \caption{Classes encountered over time in the first run NICv2-391. Each row denotes a class; colors are used to group the 50 classes in the 10 categories.  Each column denotes a training batch. A colored block in a (row, column) cell is used to indicate that at least one training session of the row class is present in the column batch. Differently from the NICv2-79 and 196 scenarios, in NICv2-391 we forced each batch to contain only patterns from one class.}
    \label{fig:nic391}
\end{figure*}

In Algorithm \ref{algo:nicv2} we report the pseudocode of the procedure used for generating the NICv2 protocol. Figures \ref{fig:nic196} and \ref{fig:nic391} show the distribution of classes in a single run of NICv2-196 and NICv2-391 respectively.

\section{Hyperparameters}
The hyperparameters used in our experiments are described in tables \ref{tab:hyper_core50} and \ref{table:renormpar}. Values reported in table \ref{tab:hyper_core50} follow the naming scheme used in \cite{maltoni2019}.
Please note that:

\begin{itemize}
    \item for AR1* we used two different learning rates: one for the CWR layer and one for the remaining part of the net. This choice can be simply explained by considering that the CWR layer update procedure in AR1* is inherited from the original CWR* strategy. We empirically observed that the overall model performance largely benefits from the use of an higher learning rate for the CWR layer.
    \item the proposed ``Weight Constraining by Learning Rate Modulation'' approach was applied to the AR1* strategy only. While this approach could be easily applied to EWC as well, in our experiments we did not change the behavior of the original EWC algorithm. This will allow for a more direct and unbiased comparison of the proposed strategies.
\end{itemize}

\begin{table}[H]
\caption{Experimental setup}
\centering
\begin{tabular}{lrrr}
\toprule
                     \multicolumn{1}{c}{\textit{Component}}      & \multicolumn{1}{c}{\textit{Model/Version}} \\ \midrule
 Operating System                         & Debian 8.3                         \\
 Docker                         & 18.06.1                         \\
 Nvidia Driver                         & 390.48 (CUDA 9.0, CuDNN 7)                         \\
 CPU                          & Intel(R) Xeon(R) CPU E5-2650                          \\ 
 GPU                          & GTX 1080 Ti (11 GB VRAM)                          \\ 
 RAM                       & 64 GB DDR3 (1600 MHz)                       \\ \bottomrule
\end{tabular}
\label{table:hardware}
\end{table}

\begin{algorithm}[H]
\captionsetup{font=small}
\begin{algorithmic}[1]
\footnotesize
\Procedure{CWR+}{}
\State $cw=0$
\State $\text{init } \bar{\Theta} \text{ random or from pre-trained model (e.g. on ImageNet)}$
\State \textbf{for each} $\text{training batch } B_i$:
\State \ \ \ \ expand output layer with neurons for the new classes in $B_i$
\State \ \ \ \ $tw=0$ (for all neurons in the output layer)
\State \ \ \ \ train the model with SGD on the $s_i$ classes of $B_i$:
\State \ \ \ \ \ \ \ \ \textbf{if} $B_i=B_1$ learn both $\bar{\Theta}$ and $tw$
\State \ \ \ \ \ \ \ \ \textbf{else} learn $tw$ while keeping $\bar{\Theta}$ fixed
\State \ \ \ \ \textbf{for each} class $j$ among the $s_i$ classes in $B_i$
\State \ \ \ \ \ \ \ \ $cw[j]=tw[j]-avg(tw)$
\State \ \ \ \ test each class $j$ by using $\bar{\Theta}$ and $cw$
\EndProcedure
\end{algorithmic}
\caption{CWR+ pseudocode for NC scenario where each training batch $B_i$ includes patterns of new classes only: $\bar{\Theta}$ are the class-shared parameters of the representation layers; the notation  $cw[j]$ / $tw[j]$ is used to denote the groups of consolidated / temporary weights corresponding to class $j$. The mean-shift in line 11 allows to adapt the scale of parameters trained in different batches (see Section 3.2 of \cite{maltoni2019}).}\label{algo:cwr+}
\end{algorithm}

\begin{algorithm*}
\captionsetup{font=small}
\begin{algorithmic}[1]
\footnotesize
\Procedure{NICv2}{num\_runs, num\_batches, max\_start}
\State $num\_runs$: number of sequences to produce. Since in continual learning the pattern presentation order has an impact on the accuracy, experiments \phantom \ \ \ \ \ \ \ \phantom \ \ \ \ \ \ \ \ \ \ \ \ \ \ \ \ \ \ \ \ \ need to be averaged over multiple runs. In this paper we used $num\_runs = 10$.
\State $num\_batches$: the total number of training batches (refer to Table 1).
\State $max\_start$: we need to limit the maximum position for the insertion point of classes to leave some room to accommodate all their training sessions.
\State \textbf{for each} run in $num\_runs$:
\State \ \ \ \ assign to $B_1$ 10 training sessions (by selecting 1 class from each category)
\State \ \ \ \ \ \ \ \ \textbf{for each} class $C$ of the ramaining 40 classes:
\State \ \ \ \ \ \ \ \ \ \ \ \ random sample $insertion\_point \in [1, max\_start]$
\State \ \ \ \ \ \ \ \ \ \ \ \ \ \ \ \ \textbf{for each} training sessions $S$ of class $C$:
\State \ \ \ \ \ \ \ \ \ \ \ \ \ \ \ \ \ \ \ \ $assigned = false$
\State \ \ \ \ \ \ \ \ \ \ \ \ \ \ \ \ \ \ \ \ \textbf{while} not $assigned$:
\State \ \ \ \ \ \ \ \ \ \ \ \ \ \ \ \ \ \ \ \ \ \ \ \ random sample current batch $B_c$ with $c \in$ [$insertion\_point, num\_batches$]
\State \ \ \ \ \ \ \ \ \ \ \ \ \ \ \ \ \ \ \ \ \ \ \ \ \textbf{if} $B_c$ is not full:
\State \ \ \ \ \ \ \ \ \ \ \ \ \ \ \ \ \ \ \ \ \ \ \ \ \ \ \ \ assign training session $S$ to $B_c$
\State \ \ \ \ \ \ \ \ \ \ \ \ \ \ \ \ \ \ \ \ \ \ \ \ \ \ \ \ $assigned = true$
\EndProcedure
\end{algorithmic}
\caption{NICv2 generation pseudocode: the first batch includes one training session of a single class within each category. Then for each of the remaining 40 classes, we randomly sample the minimum allowed insertion point and then assign all the remaining training sessions to batches in the permitted range. The size of training batches (checked at line 14) depends on num\_batches (see Table 1). Note that the algorithm might loop for unappropriated values of some parameters.}\label{algo:nicv2}
\end{algorithm*}

\begin{table}[!thb]
    \caption{Hyperparameter values used in our experiments. The selection was performed on run 0, and hyperparameters were then fixed for runs $1, \dots,9$. We used the same set of hyperparameters for the NICv2 79, 196 and 391 scenarios. In this table we use to the same notation introduced in \cite{maltoni2019}.}
  	\label{tab:hyper_core50}
  \begin{tabular}[H]{ll}
    \toprule
    \multicolumn{2}{c}{\textbf{Naive}}   \\
    \cmidrule(r){1-2}
    \multicolumn{1}{c}{\textit{Parameters}}    & \multicolumn{1}{c}{\textit{MobileNet V1}}  \\
    \midrule
    Head  							& Maximal  				     \\
    $B_1$: epochs, $\eta$ (learn. rate)   		& 2, 0.001		 		      \\
    $B_i, i>1$: epochs, $\eta$ (learn. rate)   	& 2, 0.000035    	 		  \\
    \toprule
    \multicolumn{2}{c}{\textbf{LWF}}   \\
    \cmidrule(r){1-2}
    \multicolumn{1}{c}{\textit{Parameters}}    & \multicolumn{1}{c}{\textit{MobileNet V1}}    \\
    \midrule
    Head  							& Maximal  			     \\
    $\lambda$										& 0.1  \\
    $B_1$: epochs, $\eta$ (learn. rate)   		& 2, 0.001		 		      \\
    $B_i, i>1$: epochs, $\eta$ (learn. rate)   	& 2, 0.00005 \\
    \toprule 
    &
    
    \\
    \toprule
    \multicolumn{2}{c}{
    \textbf{EWC}}   \\
    \cmidrule(r){1-2}
    \multicolumn{1}{c}{\textit{Parameters}}    & \multicolumn{1}{c}{\textit{MobileNet V1}}     \\
    \midrule
    Head 							& Maximal  			     \\
    $max_F$										& 0.001 			  \\
    $\lambda$									& 2.0e6				 \\
    $B_1$: epochs, $\eta$ (learn. rate)   		& 2, 0.001		 	      \\
    $B_i, i>1$: epochs, $\eta$ (learn. rate)   	& 2, 0.0001    	  \\
    \toprule
    \multicolumn{2}{c}{\textbf{CWR*}}   \\
    \cmidrule(r){1-2}
    \multicolumn{1}{c}{\textit{Parameters}}    & \multicolumn{1}{c}{\textit{MobileNet V1}}     \\
    \midrule
    Head 							& Maximal  			     \\
    $B_1$: epochs, $\eta$ (learn. rate)   		& 4, 0.001		 	      \\
    $B_i, i>1$: epochs, $\eta$ (learn. rate)   	& 4, 0.001      \\
    \toprule
    \multicolumn{2}{c}{\textbf{AR1}}   \\
    \cmidrule(r){1-2}
    \multicolumn{1}{c}{\textit{Parameters}}    & \multicolumn{1}{c}{\textit{MobileNet V1}}     \\
    \midrule
    Head 							& Maximal  			     \\
     			     
    $w_1, w_i (i > 1)$ 							& 0.5, 0.5  	     \\
    $max_F$										& 0.001 			  \\
    
    $B_1$: epochs, $\eta$ (learn. rate)   		& 4, 0.001		 	      \\
    $B_i, i>1$: epochs   	& 4 \\
    \ \ \ \ \ $\eta$ (learn. rate, CWR layer)   	&  0.001  
    \\
    \ \ \ \ \ $\eta$ (learn. rate,other layers)   	&  0.0001 \\
    \toprule
    \multicolumn{2}{c}{\textbf{DSLDA}}   \\
    \cmidrule(r){1-2}
    \multicolumn{1}{c}{\textit{Parameters}}    & \multicolumn{1}{c}{\textit{MobileNet V1}}     \\
    \midrule
    shrinkage   		& 1e-4		 	      \\
    sigma   	& plastic \\
    \bottomrule
  \end{tabular}
\end{table}

\begin{table}[H]
\caption{Batch ReNormalization parameters}
\centering
\begin{tabular}{p{2cm}rrr}
\toprule
                     \multicolumn{1}{c}{\textit{Parameters}}      & \multicolumn{1}{c}{\textit{NICv2-79}} & \multicolumn{1}{c}{\textit{NICv2-196}} & \multicolumn{1}{c}{\textit{NICv2-391}} \\ \midrule
$R_{max}$                       & 1.25                        & 1.25                         & 1.5                          \\ 
$D_{max}$                       & 0.5                         & 0.5                          & 2.5                          \\ 
Moving Avg. update rate & 0.9999                      & 0.9999                       & 0.9999                       \\ \bottomrule
\end{tabular}
\label{table:renormpar}
\end{table}







\end{document}